# Exemplars can Reciprocate Principal Components

Kieran Greer, Distributed Computing Systems, Belfast, UK.
kgreer@distributedcomputingsystems.co.uk.
Version 1.1

***Abstract –*** This paper presents a clustering algorithm that is an extension of the Category Trees algorithm. Category Trees is a clustering method that creates tree structures that branch on category type and not feature. The development in this paper is to consider a secondary order of clustering that is not the category to which the data row belongs, but the tree, representing a single classifier, that it is eventually clustered with. Each tree branches to store subsets of other categories, but the rows in those subsets may also be related. This paper is therefore concerned with looking at that second level of clustering between the other category subsets, to try to determine if there is any consistency over it. It is argued that Principal Components may be a related and reciprocal type of structure, and there is an even bigger question about the relation between exemplars and principal components, in general. The theory is demonstrated using the Portugal Forest Fires dataset as a case study. The Category Trees are then combined with other Self-Organising algorithms from the author and it is suggested that they all belong to the same family type, which is an Entropy-style of classifier.

## 1   Introduction

This paper presents a clustering algorithm that is an extension of the Category Trees [9] algorithm. Category Trees is a clustering method that creates tree structures that branch on category type and not feature. The development in this paper is to consider a secondary order of clustering that is not the category to which the data row belongs, but the tree, representing a single classifier, that it is eventually clustered with. Each tree branches to store subsets of other categories, but the rows in those subsets may also be related. This paper is therefore concerned with looking at that second level of clustering between the other category subsets, to try to determine if there is any consistency over it. The mathematics for this problem could get quite complex and it will be argued that Principal Components [16] may be a related and reciprocal type of structure. This therefore poses an even bigger question about the relation between exemplars and principal components, in general. However, only a lightweight analysis of the mathematical properties is possible, really at the level of matching the different ingredients that are involved. The success criterion can also be decided artificially and may be the resulting clustered column with the lowest variance, or a lower variance compared to other results. It has then been decided that a number of classifiers created by the author in fact belong to the same family type, which is an entropy-style of classifier. They can even be combined to produce small improvements, where tests on a set of benchmark datasets demonstrate this and the properties that make them similar are described.

The problem was formulated when considering geographical data that might change over time. It is therefore a time-series problem, but one where the data is placed into discrete time bands and is not continuous. The distributed nature of the input suggested that a category could be created from each input station, or sensor set. This is a bit unusual, because the classifier would then be expected to learn an average value for itself, but the problem was expanded to not consider this aspect, but the aspect of how the data rows in each individual tree and its branches may be grouped together. The branches represent subsets of data rows for other categories, where the original idea was that comparing these subsets for feature analysis would be more accurate than looking at the whole dataset. With time series, it would also be possible to consider how the relation changes over time. In fact, that is a more advanced problem than what has been concluded for this paper, where this test case considers only the relation between all rows in the branches together. The theory was tested with the forest fires dataset [2] as a specific case and to verify the result, the smaller El Nino dataset [3] has also been looked at.

The rest of this paper is organised as follows: section 2 describes some related work and section 3 summarises the Category Trees again. Section 4 describes the environmental case study and gives a result that might help when monitoring forest fires. Section 5 describes why category trees as exemplars, might reciprocate principal components and also produces new tests for combining a set of the author's own classifiers. Section 6 describes the properties of the type of classifier that they all belong to, while section 7 gives some conclusions on the work.

## 2   Related Work

The theory may have overlap with Principal Component Analysis (PCA) [16] and even the work of Oja [14] and others, who created neural network systems from these features. While principal components were learned, the problem reduced to a minimum when the features became similar, like a regression problem. It is interesting that Category Trees [9] originated from trying to oscillate an error around a mean-centred value, created from a wave shape [11] representing the input data rows. That then became simply adjusting to the averaged batch row value. More specifically, similar ideas could be:

- Zero-mean could relate to the original oscillating error. It was found however that if dealing with a single averaged batch value, a single adjustment would suffice instead.
- Then of course the averaged values themselves and the closest fit to it. Not to realise maximum difference but to align the data rows with maximum similarity.
- Variance was also a part of the theory for a new entropy equation [5] that is intended to give some measure of cohesion across a pattern. It can also measure the cohesion across a tree structure, with possibly arbitrary concepts in the branches.

Measuring the change in the variance is possibly more associated with Kullback-Leibler divergence [12] and Information Gain theory [13]. As described in Wikipedia[1], the Kullback–Leibler divergence, (also called relative entropy), is a measure of how one probability distribution is different from a second, reference probability distribution. The reference distribution would be the original category exemplar and the new distribution would be the exemplar after the data rows were re-clustered. Entropy is represented in this paper by the variance. Information Gain is then a method for measuring the entropy difference over different sets of variables. It does this by determining if the data can be split into certain subsets that may hold more consistent information than the whole dataset together. If the sum of the variances in the subsets is less than for the whole dataset, then that difference is the information gain, and it is also a reduction in the variance.

---

[1] https://en.wikipedia.org/wiki/Kullback–Leibler_divergence

## 3   Category Trees

This section is a review of the Category Trees classifier [9]. The method is supervised, where each actual output category is assigned a classifier and the classifier learns to adjust from an averaged batch row value to the desired output value. This adjustment can in fact be done in a single step. Because each category is separate, it is like a neural network with a separate neuron for each output category and the value can be anything, but would typically be 1 or 0.5, for example. The classifier weight set therefore adjusts from the averaged batch row value to the value 0.5 or 1. Because the adjustment is for the batch value, individual data rows may be closer to other classifier results. Therefore, after learning this adjustment, each data row is passed through each classifier again, and it is stored with the classifier that produces the smallest error with it. If the data row's category is not the same as the classifier, then a new layer is created in the classifier and the data row is stored with a new classifier in the new layer that represents the new category. The difference being that subsequent layers would be trained on smaller and smaller subsets of data rows and the base classifier is always representative of the original category. Then to retrieve the category information, a data row is passed to each classifier and the one with the smallest error in the base layer is used. If it has branches, then the row is passed to each branch and the one with the smallest error again used, until the process terminates at a leaf node. The category of that leaf node is then declared the category for the data row.

The process is illustrated in Figure 1. The dataset contains two categories – A and B. Two classifiers are thus trained, one to recognise the averaged category A value and one for the averaged category B value. When each individual row is then passed through the classifiers, some of the B category rows are closer to the averaged category A value. Therefore, the A classifier branches to a new layer, where it recognises both its own category A rows and a subset of the category B rows. While this process should be very accurate, it is quite a shallow architecture and it may not generalise as well as it is able to learn the original data, but probably still well-enough to be useful.

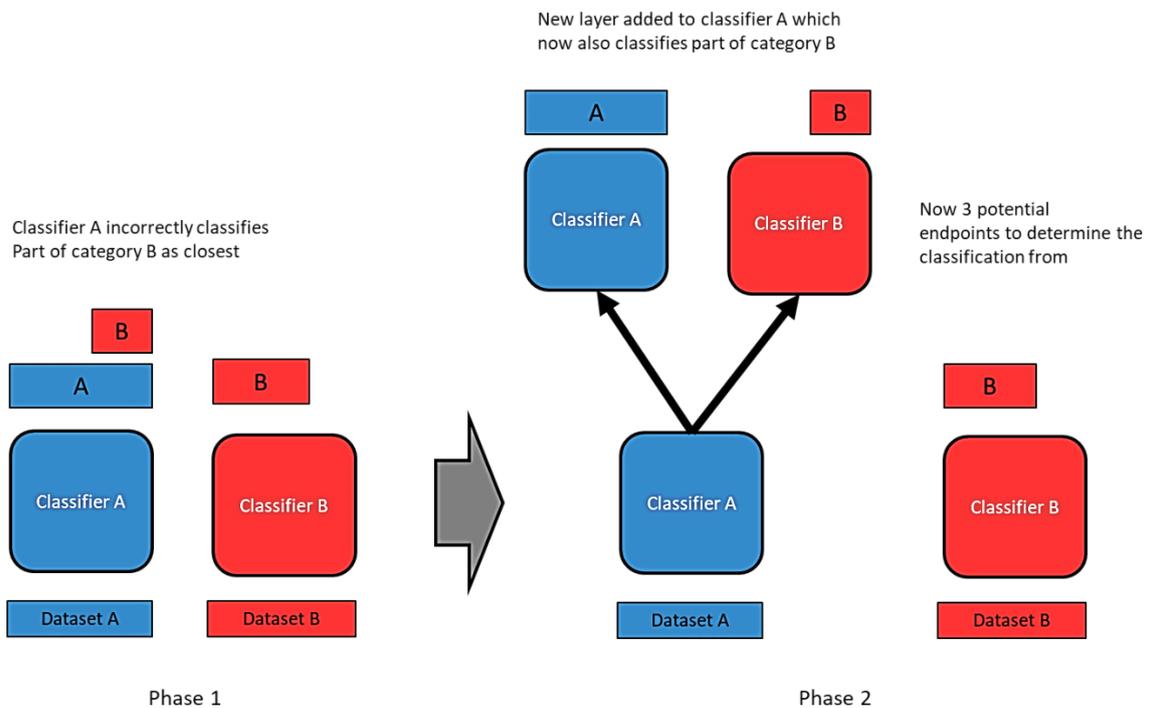

Figure 1. Schematic of the classifier in action. Phase 1 realises that classifier A also classifies part of category B better. Phase 2 then adds a new layer to classifier A, to re-classify this subset only.

### 3.1   Secondary Clustering

While each category can be distributed across any number of classifiers, there may also be the option to look at all of the data rows associated with each classifier tree separately. There is obviously some type of association there, because the data rows are closest to the classifier value and so their feature sets must also be close. And because it is for subsets of rows, the feature comparison should be more accurate than if the feature sets of the whole dataset were compared. This is one possibility with the supervised approach, when the averaged value is in fact a centroid, at the centre of the input category. The Iris dataset [4], for example, is perfectly centred, with only 10 data rows clustered with other categories. As described in section 4, another possibility may be to use the averaged value as an exemplar. In that case, it represents something, but it may not be at the centre of the category, or the category selection is artificial and so there is not as much consistency over the data rows that created it.

### 3.2 Recursive Clustering

The probability distributions created from each category are discrete, but it would be possible to create a series of updates through a recursive procedure. Quite simply, when the rows are re-clustered with the closest exemplar, the process can be repeated by creating new exemplars from the new row batches and then re-clustering on those again. This can repeat until a minimum number of changes occur. If the categories are known, then this should not improve the clustering performance by very much, but it may benefit an unsupervised or a semi-supervised procedure more. This also means that if it is better for an environmental scenario, then the distributed stations are semi-supervised at best and a fully supervised approach would have more knowledge about the correct category definitions. This therefore means that using Category Trees in a supervised environment would be an upper bound on how well it can classify a problem. Any skew is implicitly stated and the classifiers can learn it directly. If the recursive clustering can adjust to a slightly different orientation, then the hope would be that an unsupervised environment that cannot inherently learn the skew, would benefit for the adjustment more and it would help to align the guessed category set with the real ones.

## 4 Forest Fires Case Study

The forest fires dataset [2] gives readings for a year, from Portugal's Monteshino natural park in the north east of Portugal. This was actually the first dataset that was tested, when it became clear that larger datasets could not be processed easily, but it is unlikely that the solution is specific to this one dataset only. The idea is to create categories based on the distributed nature of the input and the dataset is divided into a 9x9 matrix, where each location can represent a category. The fires that occurred in each region over the period of a year were recorded with the month and day of the fire and some sensor indicators. The weather sensors recorded temperature, relative humidity, wind speed and rain. As well as this, the fire indicators of Fine Fuel Moisture Code (FFMC), Duff Moisture Code (DMC), Drought Code (DC) and Initial Spread Index (ISI) were recorded with each weather reading. The original paper tried to predict when fires might occur and was more successful with the more numerous small fires. This paper suggests some type of relation between the location cells, where if there is a fire in one grid location, then other locations might also be vulnerable. The output criterion was taken to be the month that the fire occurred in and in fact there was a clear correlation there in the results. But there was also a large spread of related grid cells and so it might not be very accurate for predicting exactly when a fire would take place.

Deciding to convert the sectors into classifiers is an arbitrary decision, albeit with some intuition. Then there is also the output criterion, which has been selected to be the month of the year. This again is arbitrary and not implicit in the data, apart from the fact that it is what is being measured. Then a subset of the other variables was used to realise the relation between these two variables. The dataset was therefore divided into baches of rows, one batch for each grid location that had fires. The classifier for that location learned the average input value and used that as an exemplar for the location. After learning the averaged value, each row was presented to each classifier and it was clustered with the one it was closest to. That produced a new set of row batches for each classifier. The month for each row in each batch was then analysed and its variance was calculated. The average variance for the rows clustered for the classifiers was compared to the variance for each grid cell without clustering and the result is shown in Table 1.

|  | Month of the Fire |
|---|---|
| Variance Before | 1.32 |
| Variance After | 0.35 |

Table 1. Variance values for each Forest Fires Month variable before and after the clustering.

The surprising result was that the classifiers aligned the rows based on the month of the fire. A description of some of the clustered rows in given in Appendix A. Further analysis showed that a classifier might not have any of its own rows clustered with it, so while the average value represents the batch set of rows, no individual row from that batch might be clustered with the classifier. As the first two clusters in Appendix A show, there are no data rows from the sector finally clustered with it. This could suggest that the clustering criterion is not distance-based. The classifier therefore does not produce a centroid, at the centre of the category, but some other representative value. This is explained further in section 5. The environmental result however, would be that because the data is clustered consistently on the month, it could help a fire service to monitor the region, as it could follow the sequences of locations when any fires started in one of them.

### 4.1   El Nino Dataset

The smaller El Nino dataset [3][1] was also tested. Any row with a missing value was removed first, resulting in 507 rows and then each buoy was allocated to be a category. This also related to a specific location and then columns 4 to 8 were used to train the classifier. The variance of each of these columns was then compared before and after the training, for each category, where the result is shown in Table 2.

|  | Zon. Wind | Mer. Wind | Humidity | Air Temp. | S.S Temp. |
|---|---|---|---|---|---|
| Var Before | 1.15 | 1.1 | 2.1 | 0.38 | 0.245 |
| Var After | 0.85 | 0.8 | 1.5 | 0.26 | 0.25 |

Table 2. Variance values for each El Nino variable before and after the clustering.

The re-clustering of the rows has therefore produced a smaller variance in 4 of the 5 variables and an almost equal variance in the fifth one. This is slightly different to the forest fires dataset, when the output variable was not included in the training rows.

## 5   Category Trees as Principal Components

The case study shows that a classifier may not end up with any of its own category rows clustered with it. If the relation was for centroids, then it would be expected that the rows for the category would be clustered with it and so possibly each classifier value is an exemplar instead. Therefore, the clustering is not based on distance alone and the use of exemplars might point to a principal component that is not aligned exactly with its input data, but with some type of feature set. Further analysis, probably based on feature selection, would be interesting. For example, why is this sector's value, created from a different set of rows and times, an exemplar for that set of rows?

Principal Component Analysis converts the data into vectors of maximum variance. The PCA vector lines replace the data rows and when learned can be used to discriminate where the dataset changes most. Principal components are statistically the most significant features. The exemplars of this paper would instead represent directions of minimum variance, or maximum similarity. While that similarity is defined by some arbitrary criterion, it may be possible to formulate this problem in a general sense, possibly along the lines of the Information Gain algorithm [13]. It could be imagined that in an unsupervised setting, where these variable sets are not known, some iterative process could try different combinations and maybe measure the variance over the output criteria, to select the best matching combinations with the lowest variances. Therefore, we may have a reciprocal situation, where the columns and differences of PCA can be compared with the rows and similarity in this method. The following sections however try a different method for some of the author's own algorithms.

### 5.1   Tests over Benchmark Datasets

Tests have been carried out over some benchmark datasets found in the UCII Machine Learning Repository [15]. As stated in section 3.2, using Category trees in a supervised environment would be an upper limit on the performance. An earlier paper [6] considered a self-organising algorithm that was able to determine a number of arbitrary clusters for itself and produced very reasonable results. The tests for that algorithm are repeated here, but then compared to a new version that also includes the Category trees. The self-organising algorithm uses a closest distance measure to cluster individual nodes and then a new Frequency Grid [8] to further cluster those events. Because the supervised Category Trees can learn an inherent skew and even re-align to a different set of vectors, using the unknown results of the self-organising algorithm as the base cluster set for the trees may help to re-align some of those data rows again, to produce a more correct orientation. Then there is also the idea of recursively feeding the Category Trees result back to itself. In this case, only 1 recursive feedback was used. The results of using these 3 algorithms are given in Table 3, where:

- SO is the self-organising algorithm by itself [6].
- SO-CT is the self-organising algorithm, that feeds into standard category trees, that feed into the next self-organising level.
- SO-SRCT is the self-organising algorithm, that feeds into category trees that use the secondary ordering and 1 phase of recursion, before feeding back into the next self-organising level.

|  | SO | SO-CT | SO-SRCT |
|---|---|---|---|
| **Iris (150-3)** | 5-94% V1 | 7-93% V1 | 4-91% V1 |
| **Wine (178-3)** | 8-91% V1 | 5-98% V1 | 9-95% V1 |
| **Zoo (101-7)** | 8-99% V2 | 15-98% V2 | 15-98% V2 |
| **Hayes-Roth (132-3)** | 4-69% V1 | 4-71% V1 | 4-77% V2 |
| **Heart Disease Cleveland (303-5)** | 7-77% V2 | 7-73% V2 | 13-80% V1 |
| **Sonar (208-2)** | 4-82% V2 | 3-88% V1 | 3-88% V2 |
| **Wheat Seeds (210-3)** | 7-88% V2 | 10-97% V1 | 8-97% V1 |
| **Average Cluster Error** | 2.44 | 3.7 | 4.3 |
| **Average Accuracy Percentage** | 85.5% | 86.5% | 89.5% |

Table 3. Test results on some benchmark datasets.

The self-organising algorithm scores very well by itself, so there is not much room for improvement. Note that each test was run over 3 iterations instead of 10 and this produced better results than what was published in [6]. The results from 50 tests were averaged and they are also for a randomised ordering of the dataset. They also showed a much more even split between version 1 or version 2 of the frequency grid. The final number of clusters can vary a bit for the same error percentage, where the final clusters were selected as the set before the next level reduces to below the desired number.

## 5.2  Test Conclusions

The results are very pleasing and are what the author would like to demonstrate. But the values can still change quite a lot, so the published results should be taken with a pinch of salt, but they are generally accurate for what each algorithm would produce. While the error is improved, the number of clusters also increases, which is not a good result, but not as serious as the error reduction. The idea is that the supervised Category Trees can introduce a skew to the self-organised clusters and that will help them to re-align to a more accurate set of exemplars. If the secondary clusters and recursive feedback also has an effect, then better still, because the first category tree learning phase can only learn the cluster set presented by the self-organising algorithm. The adjustment comes in the second recursive stage. Note however that if an optimum result is achieved, any further stage will move away from it and so only 2 stages were tried in the recursive procedure. Also, if the test dataset is properly formed, then the exemplars are also centroids and the secondary clusters would be almost the same.

Changing the number of test runs had a significant effect, where it was reduced from 10 to 3 in this set of tests. It can sometimes be the case when dealing with averaged values that a global minimum occurs when the correct function is realised. Because the averaged values are less variable, then as the values move to the minimum error points, they become more similar, when it is more difficult to discriminate over them. In that case, it is sometimes better to stop the learning process before the minimum error is achieved and keep more variability in what the classifier would produce as output. This may be why fewer test runs gives a better performance, because there is less aggregating of everything at the end of it. Overtraining is a well-known problem, but that is typically when comparing with a different test dataset and it usually still produces a better fit in the train dataset. Concerning the preference for version 2 of the frequency grid, it produces larger clusters to start with and so it would probably work better with fewer ensemble aggregations afterwards. The classifier rather than the overlap determines the clusters.

## 6   The Same Family of Classifiers

A second tree structure suggested by the author is a Concept Tree [11][7]. This stores concepts of arbitrary nature in a tree structure with the simple mathematical rule that a child node cannot have a larger frequency count than its parent. It is interesting that Category Trees also supports this rule. As classifiers in branches store only subsets of nodes, they are likely to reduce their number of rows each time, which is like reducing a frequency count. If the variance inside of each tree is less than for the global population, then that gives the desired cohesion that the Concept Tree requires. It may be the case that Category Trees and Concept Trees belong to the same family of classifier, but more than that – the Frequency Grid as well. Category Trees are maybe a supervised version, where Concept Trees are semi-supervised or unsupervised. This family of classifier makes use of averaged or batch values instead of the single local values. It is concerned with reducing the entropy or variance of the structure, which is again a global error correction over the smaller local corrections. Even the Frequency Grid clusters use the largest count from a global table and not closest distance. It may be the case that the counting rule of the Concept Tree that also occurs in the Category Tree structure leads to global minima over the much more diverse landscape that a neural network might traverse. If the search space is a global minimum, even a relatively shallow one, then it makes sense that single corrections can move to the best result and many iterations are not required. This is also interesting, because maybe in more abstract terms, operations for one of the classifiers can be tried on the other one. The similarity becomes even more pronounced when both tree structures perform a slight adjustment, through an internal recursive update, as described here in section 3.2 and in the paper [7] section 7.3, for the Concept Trees. In that paper, it was like a matching process that refined itself slightly.

## 7 Conclusions

This paper has introduced a new clustering method that can improve the variance for similarity, across the cluster sets. It is based on the Category Trees algorithm, which is a supervised approach, but the new method may eventually work in an unsupervised manner. The method creates exemplars instead of centroids and poses the question if exemplars and principal components are related in some way. Environmental data is used as the case study and the result could help a fire service to monitor a region for fires. A set of the author's algorithms have been placed into a family type that is an entropy-style of classifier. Test results on combinations of this family show improved results to what was published previously and also that the different classifiers work very well together.

## Appendix A – Examples of the Forest Fires Clusters

This appendix lists some of the clusters generated by the secondary ordering of the Category Trees classifier. There were several clusters for the August – September time period, where two are displayed first. Note that there are no rows from the (1, 2) sector in that cluster or from the (1, 3) sector in that cluster.

Rows clustered for Sector (1, 2)

X, Y, month, day, FFMC, DMC, DC, ISI, temp, RH, wind, rain, area

2,4,aug,tue,94.8,108.3,647.1,17,20.1,40,4,0,0
2,4,aug,wed,92.1,111.2,654.1,9.6,20.5,35,4,0,1.64
4,3,aug,wed,92.1,111.2,654.1,9.6,20.4,42,4.9,0,0
4,3,aug,wed,92.1,111.2,654.1,9.6,20.4,42,4.9,0,0
6,3,aug,fri,91.1,141.1,629.1,7.1,19.3,39,3.6,0,1.56
6,4,aug,thu,95.2,131.7,578.8,10.4,20.3,41,4,0,1.9

Rows Clustered for Sector (1, 3)

1,4,aug,wed,91.7,191.4,635.9,7.8,19.9,50,4,0,82.75
2,2,sep,fri,92.4,117.9,668,12.2,23,37,4.5,0,0
2,2,aug,tue,92.1,152.6,658.2,14.3,21.8,56,3.1,0,0.52
3,4,sep,sun,92.4,124.1,680.7,8.5,22.5,42,5.4,0,0
4,4,sep,sun,93.5,149.3,728.6,8.1,22.9,39,4.9,0,48.55
5,4,aug,tue,88.8,147.3,614.5,9,17.3,43,4.5,0,0
5,4,aug,tue,95.1,141.3,605.8,17.7,24.1,43,6.3,0,2
6,5,sep,fri,93.3,141.2,713.9,13.9,22.9,44,5.4,0,0
7,4,aug,sun,91.4,142.4,601.4,10.6,20.1,39,5.4,0,2.74
7,5,aug,tue,96.1,181.1,671.2,14.3,21.6,65,4.9,0.8,0
8,6,aug,tue,92.1,152.6,658.2,14.3,20.1,58,4.5,0,9.27
8,6,aug,tue,96.1,181.1,671.2,14.3,21.6,65,4.9,0.8,0

These were different cluster sets for the Winter period:

Rows Clustered for Sector (4, 6)

3,5,mar,mon,87.6,52.2,103.8,5,9,49,2.2,0,0
4,5,jan,sun,18.7,1.1,171.4,0,5.2,100,0.9,0,0
4,6,dec,sun,84.4,27.2,353.5,6.8,4.8,57,8.5,0,8.98
4,6,dec,thu,84.6,26.4,352,2,5.1,61,4.9,0,5.38
4,6,dec,fri,84.7,26.7,352.6,4.1,2.2,59,4.9,0,9.27
6,3,feb,sun,84.9,27.5,353.5,3.4,4.2,51,4,0,0
6,3,nov,tue,79.5,3,106.7,1.1,11.8,31,4.5,0,0
6,5,jun,sat,53.4,71,233.8,0.4,10.6,90,2.7,0,0
8,6,dec,wed,84,27.8,354.6,5.3,5.1,61,8,0,11.19

Rows Clustered for Sector (7, 3)

3,4,dec,mon,85.4,25.4,349.7,2.6,4.6,21,8.5,0,10.73
4,4,dec,mon,85.4,25.4,349.7,2.6,4.6,21,8.5,0,17.85
4,4,dec,mon,85.4,25.4,349.7,2.6,4.6,21,8.5,0,22.03
4,4,dec,mon,85.4,25.4,349.7,2.6,4.6,21,8.5,0,9.77
6,5,dec,tue,85.4,25.4,349.7,2.6,5.1,24,8.5,0,24.77